%% file: main.tex
\documentclass[conference]{IEEEtran}
\IEEEoverridecommandlockouts
\usepackage[T1]{fontenc}
\usepackage[cmintegrals]{newtxmath}
\usepackage{comment}
\usepackage{algorithmic}
\usepackage{algorithm}
\usepackage{graphicx}
\usepackage{amsmath}
\usepackage[subrefformat=parens]{subcaption}
\usepackage{booktabs}
\usepackage{cite}
\usepackage {flushend}

\usepackage{stfloats} 
\usepackage{makecell} 

\usepackage{mathtools}
\usepackage{optidef}
\usepackage{multirow}
\usepackage{cuted}

\usepackage{bbm}

\DeclareMathOperator*{\argmax}{arg\,max}

\usepackage[top=0.7in,bottom=1.0in,left=0.6in,right=0.6in]{geometry}
\def\BibTeX{{\rm B\kern-.05em{\sc i\kern-.025em b}\kern-.08em
    T\kern-.1667em\lower.7ex\hbox{E}\kern-.125emX}} 

\begin{document}
\bstctlcite{IEEEexample:BSTcontrol}
\setlength\floatsep{5pt}
\setlength\textfloatsep{5pt}
\setlength\intextsep{0pt}
\setlength\abovecaptionskip{0pt}
\title{Load-Aware Training Scheduling for Model Circulation-based Decentralized Federated Learning
\thanks{This work was supported in part by JSPS KAKENHI Grant Number JP23H00464, JST PRESTO Grant Number JPMJPR2035.}
}

\setcounter{topnumber}{5}
\def\topfraction{1.00}
\setcounter{bottomnumber}{5}
\def\bottomfraction{1.00}
\setcounter{totalnumber}{10}
\def\textfraction{0.04}
\def\floatpagefraction{0.7}

\author{
  \IEEEauthorblockN{
    \normalsize Haruki~Kainuma\IEEEauthorrefmark{1},
    \normalsize Takayuki~Nishio\IEEEauthorrefmark{1}
  }
  \IEEEauthorblockA{
    \IEEEauthorrefmark{1}\small School of Engineering, Institute of Science Tokyo, Meguro, Tokyo, Japan. E-mail: nishio@ict.eng.isct.ac.jp\\
  }
}

\maketitle

\input{0_abstract.tex}


\IEEEpeerreviewmaketitle

\input{1_introduction.tex}

\input{2_proposed_method.tex}

\input{3_experimental_evaluation}

\input{4_conclusion}

\bibliographystyle{IEEEtran}
\bibliography{main}

\end{document}

%% file: 0_abstract.tex
\begin{abstract}
This paper proposes Load-aware Tram-FL, an extension of Tram-FL that introduces a training scheduling mechanism to minimize total training time in decentralized federated learning by accounting for both computational and communication loads. The scheduling problem is formulated as a global optimization task, which—though intractable in its original form—is made solvable by decomposing it into node-wise subproblems. To promote balanced data utilization under non-IID distributions, a variance constraint is introduced, while the overall training latency, including both computation and communication costs, is minimized through the objective function. Simulation results on MNIST and CIFAR-10 demonstrate that Load-aware Tram-FL significantly reduces training time and accelerates convergence compared to baseline methods.

\end{abstract}

%% file: 1_introduction.tex
\section{Introduction}

Federated learning (FL) enables model training without exporting data, making it particularly effective for privacy-sensitive applications. In particular, decentralized FL, which can be executed solely among nodes that retain data, eliminates the need for a parameter server—a potential single point of failure—and thus enhances robustness, attracting significant attention. This approach is especially promising for cross-silo federated learning, where FL is performed among a small number of trusted data silos such as hospitals, banks, and research institutions.

Various methods have been proposed to realize decentralized FL. One representative approach is distributed SGD (stochastic gradient descent).
In Gossip SGD, each node asynchronously exchanges parameters with randomly selected peers and updates the model through aggregation \cite{blot2016gossip, jin2016scale, ormandi2013gossip}. However, Gossip SGD suffers from performance degradation due to convergence to local optima when the label distribution of training data is non-IID (Non-Independent and Identically Distributed) \cite{niwa2020edge}.
PDMM SGD addresses this issue by incorporating constraints that reduce the model differences between nodes during updates \cite{niwa2020edge}. Nevertheless, under strongly non-IID environments, it remains difficult to improve global model accuracy \cite{Tram-FL}.
CMFD mitigates this by exchanging logits instead of parameters among nodes and performs distillation on public data to optimize the output function \cite{Taya_2022}. However, it requires the availability of public data, which poses a limitation. Additionally, these DFL methods involve frequent communication, resulting in increased overall network traffic.



\begin{figure}[t!]
    \centering
    \begin{minipage}{0.3\columnwidth}
        \includegraphics[width=0.9\linewidth]{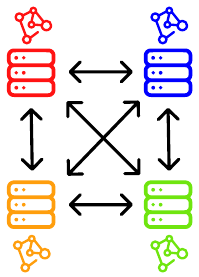}
        \subcaption{Conventional \newline DFL}
        \label{fig:conventional-dfl}
    \end{minipage}
    \begin{minipage}{0.3\columnwidth}
        \includegraphics[width=0.9\linewidth]{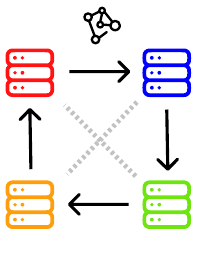}
        \subcaption{Tram-FL \newline}
        \label{fig:tram-fl}
    \end{minipage}
    \begin{minipage}{0.3\columnwidth}
        \includegraphics[width=0.9\linewidth]{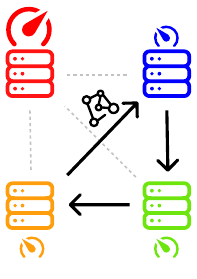}
        \subcaption{Load-aware \newline Tram-FL}
        \label{fig:latfl}
    \end{minipage}
    \caption{Comparison of DFL methods}
    \label{fig:dfls}
\end{figure}

To overcome the challenges of communication cost and non-IID data, Tram-FL has been proposed~\cite{Tram-FL}. Unlike conventional DFL methods, which maintain separate models on each node as illustrated in Fig.~\ref{fig:conventional-dfl}, Tram-FL reduces communication cost by sharing a single global model across all nodes in a cyclic manner, as shown in Fig.~\ref{fig:tram-fl}. By reducing the number of maintained models, Tram-FL achieves a substantial reduction in communication overhead at the expense of parallel training, as only one node trains the model at a time. Additionally, it performs training considering the data distribution across nodes to achieve balanced label coverage. However, a key remaining issue is that Tram-FL does not incorporate scheduling strategies that account for differences in communication bandwidth or computational capacity across nodes, limiting its efficiency under heterogeneous system conditions.

In this paper, we propose Load-aware Tram-FL, an extension of Tram-FL that improves training scheduling by explicitly considering computational and communication loads. We formulate the scheduling problem as a global optimization task that determines both the training node and the amount of data to process at each round, based on system resources and label distribution (Fig.~\ref{fig:latfl}). By decomposing this problem into node-wise subproblems, we make it solvable in practice. This load-aware scheduling enables the system to avoid nodes and communication paths under heavy load while ensuring balanced label utilization, resulting in faster convergence and reduced training time under heterogeneous conditions.

The main contributions of this paper are: (i) formulating load-aware training scheduling for Tram-FL as a global optimization problem and making it solvable by decomposition into node-wise subproblems; and (ii) experimentally demonstrating that Load-aware Tram-FL significantly reduces total training time and accelerates convergence compared to baseline strategies under various federated learning scenarios using MNIST and CIFAR-10 datasets.



%% file: 2_proposed_method.tex
\section{Proposed Method}

\subsection{System model}

This study considers a cross-silo federated learning (FL) setting, assuming collaboration among trusted entities such as hospitals or factories. A small number of participants, referred to as nodes, train local models on private data and exchange parameters for FL~\cite{kholod2020open}. We follow the standard FL assumption that each node holds labeled data; however, the data distribution is biased and non-IID, meaning certain labels may be prevalent on one node while absent on others. Since all nodes are trusted, we assume they can share statistical information about their datasets, particularly the number of samples per label, while the data samples themselves remain private and cannot be shared.

We consider an environment in which the nodes communicate with one another over a network, allowing model transmission and reception between any pair of nodes. Although the communication bandwidth between nodes is assumed to be stable, it differs across links. Furthermore, since each node concurrently engages in background communication for other purposes, the bandwidth available for FL is assumed to fluctuate. Specifically, if \( b_{i,j} \) denotes the baseline bandwidth between nodes \( i \) and \( j \), the effective bandwidth available for FL in training round \( k \) is given by  
$b^{k}_{i,j} = b_{i,j} \cdot u^{k}_{i,j},$
where \( u^{k}_{i,j} \in [0,1] \) represents the proportion of usable bandwidth.

Similarly, the computational capacities of nodes are heterogeneous and may vary across training rounds due to competing background tasks. Thus, the computational resource available to node \( i \) in round \( k \) is modeled as  
$r^{k}_i = r_{i} \cdot u^{k}_{i},$  
where \( r_i \) is the baseline computation capacity of node \( i \), and \( u^{k}_{i} \in [0,1] \) indicates the availability ratio of computational resources for FL in that round. In this paper, we assume that \(r_i\) is the ideal FLOPS of GPU.




For simplicity, we assume that the average available computational resources and communication bandwidth remain approximately constant within each round. Although these quantities may fluctuate over very short timescales, such variations are negligible compared to the duration required for a single training or communication step. We assume an ideal scenario in which each node can accurately predict its own load for the current round \cite{10.1371/journal.pone.0191939, sosonkina2022runtime}. Scheduling under a more realistic scenario, where the predicted and actual load differ probabilistically, is left as future work.

\subsection{Load-aware Tram-FL}

This section presents the core idea of Load-aware Tram-FL, a training framework that improves the node selection strategy in Tram-FL.
Therefore, the fundamental training procedure follows that of the original Tram-FL~\cite{Tram-FL}.


Algorithm~1 provides the protocol of Load‑aware Tram‑FL. As in Tram‑FL~\cite{Tram-FL}, Load‑aware Tram‑FL maintains a single global model that is transmitted among the nodes while training proceeds. In Step~1 of each round, each node predicts its computation and communication load for the current round based on its past and present load states. The node currently holding the model then collects the predicted available resource information from all nodes for that round.
In Step~2, based on the computation resources and bandwidth information obtained in Step 1 and on the prior–known label distributions, the node determines which node will perform model training in the current round and how much data it should use. The training scheduling strategy is detailed in Section~\ref{optim}. The set of nodes and the node selected in round \(k\) are denoted by \(\mathcal{I}\) and $i^k$. The term \(l_{i,c}\) denotes the number of data samples with label \(c\) held by node \(i\), and \(\mathcal{C}\) represents the set of all labels included in the training data. The term \(x_{i,c}^k\) denotes the proportion of label \(c\) data used for training in round \(k\) by node \(i\). If the assigned data amount \(l_{i^k,c}x_{i^k, c}^{k}\) is zero—i.e., if none of the nodes are deemed suitable for model updates due to poor computational or communication conditions, or to avoid label distribution bias in training—then no model transmission or training is performed. The system instead waits for a fixed interval before returning to Step~1. 
In Step~3 the model is transmitted to the node chosen in Step~2. In Step~4 the selected node trains the model using the number of data samples assigned in Step~2. We repeat Steps~1–4 for \(K\) rounds to improve model accuracy. The key difference from Tram-FL is that the training scheduling takes into account each node’s current computation and communication load. Hence, our approach is termed Load-aware.


The training scheduling problem involves optimizing the order of training and the amount of data used at each node. Ideally, we would perform a global optimization of the training schedule up to round \( K \). However, the resources available at each node vary over time. Therefore, instead of global joint optimization over all rounds, we adopt a greedy approach that selects a promising node and data amount at each round based on current system conditions.

\begin{figure}[tbp]
\begin{algorithm}[H]
    \caption{Load-aware Tram-FL}
    \label{alg1}
    \begin{algorithmic}[1]
    \STATE Initialization of model parameter $w^{0}$
    \STATE Locate $w^{0}$ to randomly selected node $i^0 \in \mathcal I$
    \FOR{each round $k = 1, 2, 3,\dots, K$}
        \STATE $\triangleright$ Step 1: Resource Information Acquisition
        \STATE Collect information about amount of available computing and communication resources in round \(k\) from all nodes
        \STATE $\triangleright$ Step 2: 
        Training Scheduling
        \STATE Decide the node $i^k \in \mathcal I$ and proportion of training data used for training \(x_{i^k,c}^k\)
        \IF{\(l_{i^k,c} x_{i^k, c}^k = 0,\ \forall c \in \mathcal{C} \)}
        \STATE Skip this round because there are no suitable node, and wait constant time 
        \STATE \textbf{continue}
        \ENDIF
        \STATE $\triangleright$ Step 3: Model Transmission
        \STATE Transmit model weight $w^{k-1}$ to node $i^k$
        \STATE $\triangleright$ Step 4: Model Update on Node \(i^k\)
        \STATE $w^{k} \leftarrow w^{k-1} - \eta \nabla F(w^{k-1})$
    \ENDFOR
    \end{algorithmic}
\end{algorithm}
\end{figure}

\subsection{Training Scheduling Problem}
\label{optim}

\def\Total#1{\mathop{\mathrm{T}^ { #1 }_\mathrm{total}}\nolimits}
\def\Round#1{\mathop{\mathrm{T}^ { #1 }_\mathrm{round}}\nolimits}
\def\Comp#1{\mathop{\mathrm{T}^ { #1 }_\mathrm{comp}}\nolimits}
\def\Comm#1{\mathop{\mathrm{T}^ { #1 }_\mathrm{comm}}\nolimits}
\def\Idle#1{\mathop{\mathrm{T}^ { #1 }_\mathrm{idle}}\nolimits}
\def\DataTotal#1{\mathop{\mathrm{S}^ { #1 }_\mathrm{total}}\nolimits}
\def\DataRound#1{\mathop{\mathrm{S}^ { #1 }_\mathrm{round}}\nolimits}

This section describes the proposed training scheduling algorithm, which is based on a greedy approach that determines the next node and data amount in each round based on predicted resource availability for the upcoming round.

This study aims to efficiently accelerate training within minimal time. To this end, the system must train a larger volume of data with minimal bias in a shorter duration. To capture this goal directly, we formulate the optimization problem \textbf{P1} in \eqref{P1} based on the desired training outcome. The objective function, defined as the ratio of the number of training samples used to the total time consumed, quantifies training efficiency; a higher value indicates more data trained in less time. The constraints include: (1) the variance of the number of samples per label used in training must remain below a predefined threshold, (2) the proportion of data used for training must lie within the interval $[0,1]$, and (3) only one node is permitted to perform training in each round. While \textbf{P1} intuitively expresses the scheduling objective, it is inherently difficult to solve directly due to its formulation. The components of problem \textbf{P1} are explained in detail below.

First, we describe the learning data-related components, corresponding to the numerator of the objective function \eqref{P1:O}, the first constraint \eqref{P1:const1}, and the third constraint \eqref{P1:const3}. The vectors $L_i$ and $X_i^k$ are defined as:
$$
L_{i} = \{\, l_{i,c} \mid c \in \mathcal{C} \,\},\quad X_{i}^k = \{\, x_{i,c}^k \mid c \in \mathcal{C} \,\}
$$
The inner product $L_i \cdot X_i^k$ represents the number of training samples used by node $i$ in round $k$. Thus, the numerator of the objective function indicates the total number of training samples used from round 1 to round $K$.

The first constraint \eqref{P1:const1} aims to keep the number of training samples balanced across labels by requiring that the variance in the number of samples per label remains below a threshold $V$. To express this, the variable $\mu$ is defined as the average number of training samples per label used from round 1 to round $K$:
$$
\mu = \frac {1} {\lvert \mathcal{C} \rvert} \sum_{c \in \mathcal{C}} \sum_{k=1}^{K} l_{i^k, c} x_{i^k, c}^k
$$
The constraint evaluates the variance relative to $\mu$ and enforces that it does not exceed $V$.

The third constraint \eqref{P1:const3} ensures that only one node performs training in each round. Since $L_i \cdot X_i^k$ represents the number of samples used by node $i$ in round $k$, the constraint restricts the number of nodes for which this value is at least 1 to no more than one. Because training may be skipped, the condition is formulated as “at most one” rather than strictly equal.

Next, we describe the load-related components, corresponding to the denominator of the objective function $O_{1}$ in \eqref{P1:O}. The term $\Round{k}$ in the denominator represents the sum of computation time, communication time, and idle time in round $k$, as defined in \eqref{round}. The individual terms $\Comp{k}$, $\Comm{k}$, and $\Idle{k}$ denote the computation, communication, and idle times per round, respectively, defined in \eqref{eq:comp}, \eqref{eq:comm}, and \eqref{eq:idle}.

The computation time $\Comp{k}$ is computed by multiplying the number of training samples $L_{i^k} \cdot X_{i^k}^k$ by the FLOPs per sample $M$ and dividing by the available computational resource $r_{i^k}^k$ (in FLOPS).
The communication time $\Comm{k}$ is calculated by dividing the model size $D_M$ (in bits) by the available bandwidth $b_{i^k,i^{k-1}}^k$; if the sending and receiving nodes are the same, communication time is treated as zero.
The idle time $\Idle{k}$ represents a fixed waiting time $H$ when no training occurs in Step~~2 of Algorithm~~1.

\begin{maxi!}|s|[2]
    {\substack{\left(X_{i}^{k}\right)_{\substack{\forall i \in \mathcal{I}, \\ \forall k \in [1,K]}}}}
    {
        O_{1}
        = \frac{\sum_{k=1}^{K} L_{i^k} \cdot X_{i^k}^k}{\sum_{k=1}^{K} \Round{k} + 1} \label{P1:O}
    }
    {\label{P1}}
    {\textbf{P1}:}
    \addConstraint{\frac 1 {\lvert \mathcal{C} \rvert} \sum_{c \in \mathcal{C}} \left( \sum_{k=1}^{K} l_{i^k,c} x_{i^k,c}^k - \mu \right)^2}{\le V \label{P1:const1}}
    \addConstraint{0 \le x_{i, c}^{k}}{\le 1,\ \forall c \in \mathcal{C},\ i \in \mathcal{I},\ k \in [1,K]}
    \addConstraint{\sum_{i \in \mathcal{I}} \mathbbm{1} \left( L_{i} \cdot X_{i}^k \ge 1 \right)}{\le 1,\ \forall k \in [1,K] \label{P1:const3}}
\end{maxi!}

\begin{align}
    \Round{k} &= \Comp{k} + \Comm{k} + \Idle{k} \label{round} \\
    \Comp{k} &=  \frac {M L_{i^k} \cdot {X_{i^k}^k}} {r_{i^k}^k} \label{eq:comp}\\
    \Comm{k} &= \frac {D_{M}} {b_{i^k,i^{k-1}}^k} \cdot \mathbbm{1} \left(i^k \neq i^{k-1}\right) \label{eq:comm} \\ 
    \Idle{k} &= H \cdot \mathbbm{1} \left( L_{i^k} \cdot {X_{i^{k}}^k} < 1 \right) \label{eq:idle}
\end{align}

Solving problem \textbf{P1} exactly requires knowledge of the computational and communication load states of all nodes far into the future, making it computationally intractable. To address this, we adopt a strategy that assumes the load in the immediate next round can be reasonably predicted, and determine the node and the amount of training data on a round-by-round basis.

From here, we focus on selecting the optimal node and determining the corresponding training data amount in a specific round \(k'\). Events from rounds 1 to \(k'-1\) are treated as fixed historical data.
Based on this idea, a new optimization problem \textbf{P2} is defined in \eqref{P2}. For notational simplicity, we define \(\DataTotal{K}\) as the total number of training samples used from round 1 to round \(K\), \(\Total{K}\) as the total elapsed time over the same rounds, and \(i'\) as the node selected in the round \(k'\). 
\[
    \DataTotal{K} = \sum_{k=1}^{K} L_{i^k} \cdot X_{i^k}^ k,\ \Total{K} = \sum_{k=1}^{K} \Round{k}, \ i' = i^{k'}
\]
Problem \textbf{P2} modifies the objective function of \textbf{P1} by replacing its variable with the training data proportions \({(X_{i}^{k'})}_{\forall i \in \mathcal I}\) in round \(k'\).


\begin{maxi!}|s|[2]
    {\substack{{(X_{i}^{k'})}_{\forall i \in \mathcal I}}}
    {
        O_{2}
         = \frac
         {L_{i'} \cdot X_{i'}^{k'} + \DataTotal{k'-1}}
        {\Round{k'} + \Total{k'-1} + 1}\label{P2:O}
    }
    {\label{P2}}
    {\textbf{P2}:}
    \addConstraint{\frac 1 {\lvert \mathcal{C} \rvert} \sum_{c \in \mathcal{C}} \left( \sum_{k=1}^{k'} l_{i^k,c} x_{i^k,c}^k - \mu \right)^2}{\le V}
    \addConstraint{0 \le x_{i, c}^{k'}} {\le 1,\ \forall c \in \mathcal{C},\ i \in \mathcal{I}}
    \addConstraint{\sum_{i \in \mathcal{I}} \mathbbm{1} \left( L_{i} \cdot X_{i}^{k'} \ge 1 \right)}{\le 1 \label{P2:const3}}
\end{maxi!}

The third constraint of this problem, \eqref{P2:const3}, is discrete in nature and therefore difficult to handle analytically. Since this constraint ensures that only one node performs training per round, we adopt a strategy wherein a node \(i'\) is selected from the set \(\mathcal I\), the objective function \eqref{P2:O} is maximized for each candidate, and the node yielding the highest \(O_{2}\) is finally adopted as the node \(i'\). We define the optimization problem \textbf{P3} in \eqref{P3} by selecting a specific node \(i'\).
\begin{maxi!}|s|[2]
    {\substack{X_{i'}^{k'}}}
    {
        O_{3}
        = \frac
        {L_{i'} \cdot X_{i'}^{k'} + \DataTotal{k'-1}}
        {\Round{k'} + \Total{k'-1} + 1}\label{P3:O}
    }
    {\label{P3}}
    {\textbf{P3}:}
    \addConstraint{\frac{1}{\lvert \mathcal{C} \rvert} \sum_{c \in \mathcal{C}} \left( \sum_{k=1}^{k'} l_{i^{k},c} x_{i^{k},c}^k - \mu \right)^2}{\le V}
    \addConstraint{0 \le x_{i', c}^{k'}}{\le 1,\ \forall c \in \mathcal{C}}
\end{maxi!}

The objective function \eqref{P3:O}, separated into the component for round \(k'\) and the cumulative contributions from previous rounds, can be rewritten by substituting \(\mathrm{S}^{k'}_{\mathrm{round}} \coloneqq L_{i'} \cdot X_{i'}^{k'}\) as:
\begin{align}
&O_{3}\left( \DataRound{k'} \right) \notag
= \frac{\DataRound{k'} + \DataTotal{k'-1}} {\frac{M \DataRound{k'}}{r_{i'}^{k'}} + \Comm{{k'}} + \Idle{{k'}} + \Total{k'-1} + 1} \notag \\
=& \frac{r_{i'}^{k'}}{M} \cdot \frac{ \DataRound{k'}  + \DataTotal{k'-1}} { \DataRound{k'} + \frac{r_{i'}^{k'}}{M} \left( \Comm{{k'}} + \Idle{k'} + \Total{k'-1} + 1\right)} \label{eq:O_LX}
\end{align}

\noindent We denote this maximum value of \(\DataRound{k'}\), which satisfies the constraints of \textbf{P3}, as \({{\mathrm{S}}^{k'}}^{\star}\).
Within the range \(1 \le \DataRound{k'}\), \(O_{3}\) increases monotonically when \(\DataTotal{k'-1} \le \frac{r_{i'}^{{k'}}}{M} \left(\Comm{{k'}} + \Idle{k'} + \Total{k'-1} + 1\right)\), and decreases otherwise. For simplicity, we treat the range \(0 \le \DataRound{k'} \le 1\) similarly, because the number of data used for training is zero in this range. Under these assumptions, \(O_{3}\) is either monotonically increasing or decreasing. Therefore, to solve \textbf{P3}, we can simply compare \(O_{3}(0)\) and \(O_{3}({{\mathrm{S}}^{k'}}^{\star})\), and adopt the \(\DataRound{k'}\) corresponding to the larger value.
Therefore, \(O_{3}\) achieves its maximum either when \(\DataRound{k'} = 0\) or \(O_{3}({{\mathrm{S}}^{k'}}^{\star})\). The relationship between the maximum of \(O_{3}\) and \(X_{i'}^{k'}\) is expressed as in \eqref{eq:OkArgmax}.

\begin{align}
    \argmax \left({O_{3}\left( \DataRound{k'}  \right)}\right)
    = \begin{cases}
        {{\mathrm{S}}^{k'}}^{\star} & O_{3}\left({{\mathrm{S}}^{k'}}^{\star}\right) \ge O_{3}\left(0\right) \\
        0 & \text{otherwise}.
    \end{cases} \label{eq:OkArgmax}
\end{align}

To obtain \({{\mathrm{S}}^{k'}}^{\star}\), we define the optimization problem \textbf{P4} as in \eqref{P4}. This problem shares the same constraints as problem \textbf{P3}, but the objective function is to maximize \(L_{i'} \cdot X_{i'}^{k'}\).

\begin{maxi!}|s|[2]
    {\substack{X_{i'}^{k'}}}
    {L_{i'} \cdot X_{i'}^ {k'} = \sum_{c \in \mathcal{C}} l_{i',c} x_{i', c}^{k'}}
    {\label{P4}}
    {\textbf{P4}: {{\mathrm{S}}^{k'}}^{\star}=}
    \addConstraint{\frac 1 {\lvert \mathcal{C} \rvert} \sum_{c \in \mathcal{C}} \left( \sum_{k=1}^{{k'}} l_{i^k,c} x_{i^k,c}^k - \mu \right)^2}{\le V \label{P4:const1}}
    \addConstraint{0 \le x_{i', c}^{k'} }{\le 1,\ \forall c \in \mathcal{C}}
\end{maxi!}

\noindent The first constraint of problem \textbf{P4} \eqref{P4:const1}, when interpreted as a function of \(x_{i',c}^{{k'}}\), becomes a quadratic inequality involving \(\lvert \mathcal{C} \rvert\) variables. Such problems can be effectively solved using optimization solvers such as COBYQA \cite{rago_thesis}.

Finally, we describe the specific procedure for determining $i^k$ and $x_{i,c}^k$ in Step~2 of Algorithm~1. For each node $i \in \mathcal I$, we temporarily set $i^k = i$ and solve \textbf{P4} to obtain ${{\mathrm{S}}^{k'}}^{\star}$ and \(X_{i'}^{k'}\). Then, we compare $O_3(0)$ and $O_3({{\mathrm{S}}^{k'}}^{\star})$, and adopt the greater value as the maximum for $O_3$ as noted in \eqref{eq:OkArgmax}. Finally, we compare the resulting maximum $O_3$ values across all $i$, and select the node and corresponding $X_{i^k}^k$ that yield the highest value.

%% file: 3_experimental_evaluation.tex
\section{Experimental Evaluation}

\def\Total#1{\mathop{\mathrm{T}^ { #1 }_\mathrm{total}}\nolimits}

\subsection{Experimental setup}

\noindent\textbf{Simulation Scenario:}\quad We conducted distributed training simulations with 3 to 10 FL nodes simulated on a single GPU server. The computational capacity \(r_i\) of each node was fixed at 10\,TFLOPS, corresponding to mid- to low-end GPUs. This capacity remained constant throughout the experiments. The proportion of available computational resources in each round, \(u_i^k\), varied per round and was randomly sampled from a uniform distribution over the closed interval \([0.01, 1]\).




Similarly, the communication capacity \(b_{i,j}\) between any two nodes \(i\) and \(j\) was set to 200\,Mbps, approximating typical Internet bandwidth. This capacity also remained fixed throughout the experiments. The proportion of available communication bandwidth per round, \(u_{i,j}^k\), was sampled independently for each round from a uniform distribution over the interval \([0.005, 1]\).

The computation time, communication time, idle time, and total training time in the simulation were numerically calculated based on the aforementioned computational and communication resources, as well as the model's data sizes and required FLOPs, which will be detailed below. These calculations follow Eqs.~\eqref{round}--\eqref{eq:idle}.

\begin{figure}
    \begin{tabular}{cc}
    \begin{minipage}{0.43\columnwidth}
        \centering
        \includegraphics[width=\columnwidth]{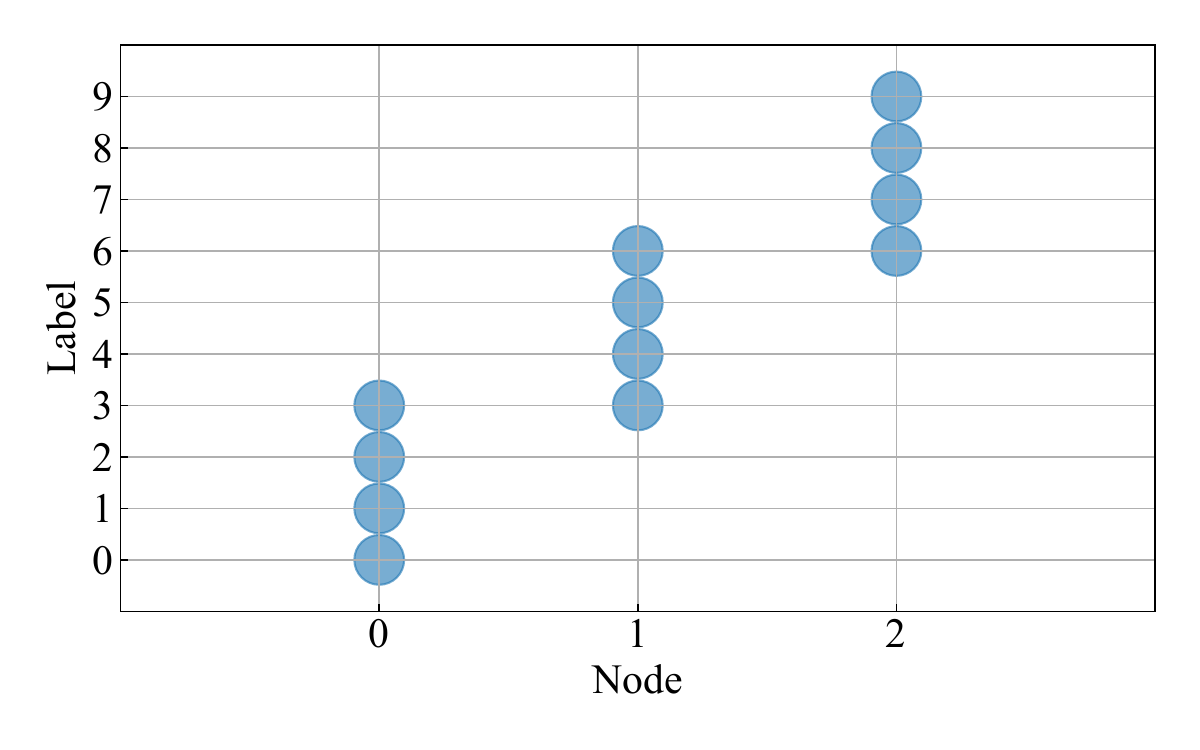}
        \subcaption{3 nodes}
    \end{minipage}
    &
    \begin{minipage}{0.43\columnwidth}
        \centering
        \includegraphics[width=\columnwidth]{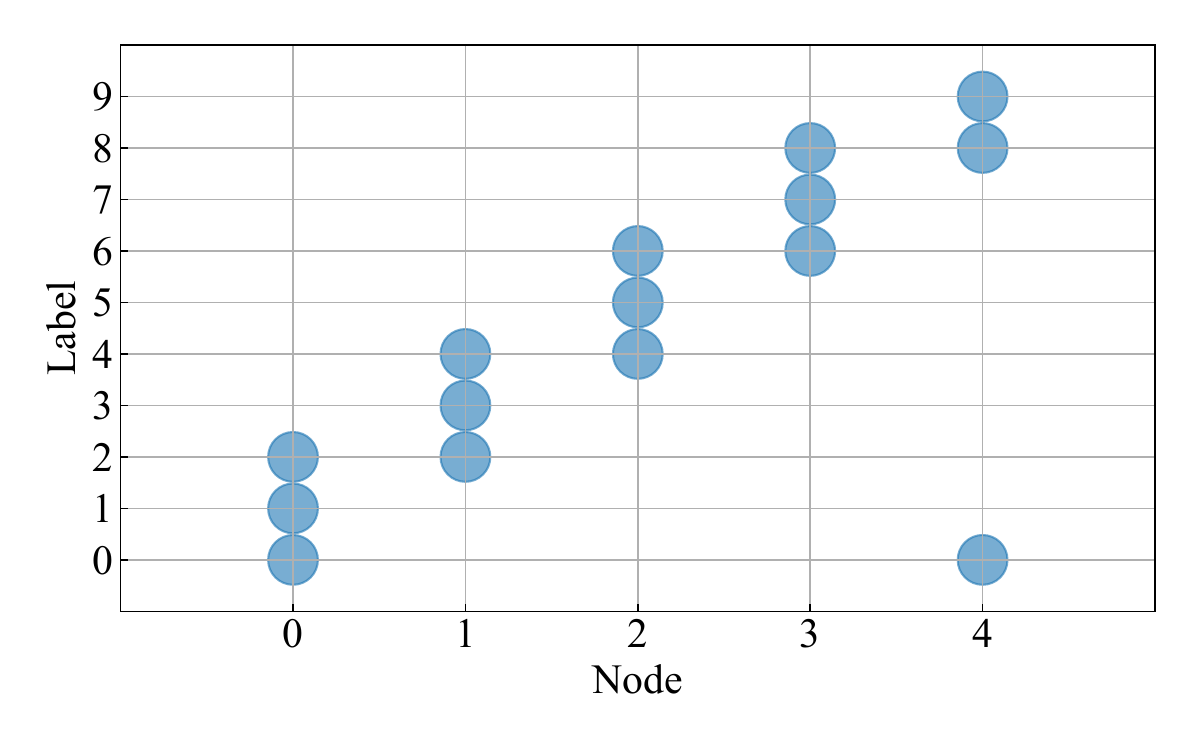}
        \subcaption{5 nodes}
    \end{minipage}
    \\
    \begin{minipage}{0.43\columnwidth}
        \centering
        \includegraphics[width=\columnwidth]{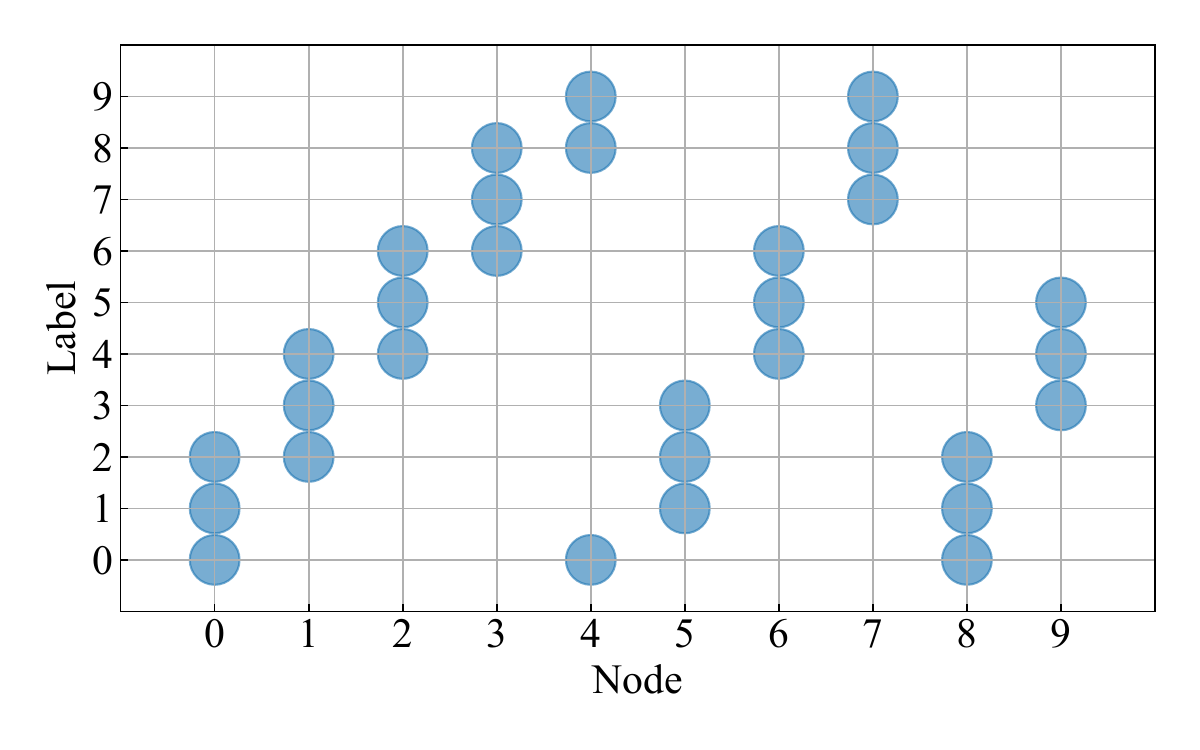}
        \subcaption{10 nodes}
    \end{minipage}
    &
    \begin{minipage}{0.43\columnwidth}
        \centering
        \includegraphics[width=\columnwidth]{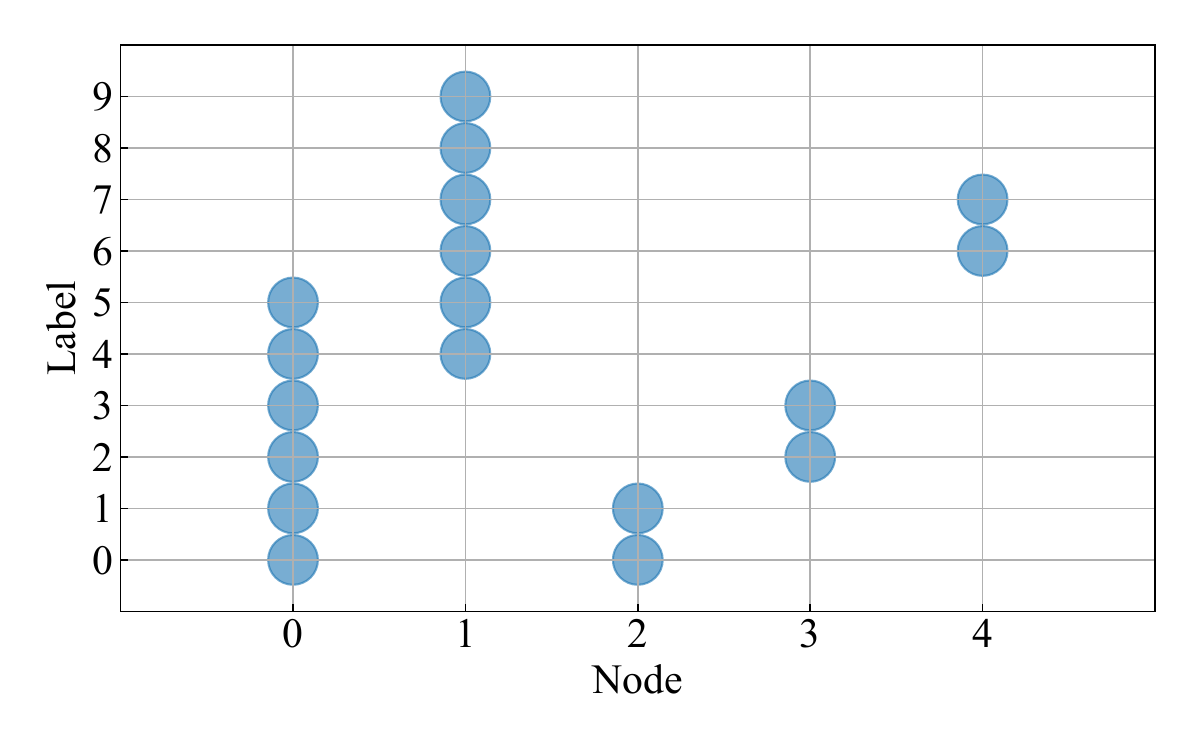}
        \subcaption{5 nodes (uneven)}
    \end{minipage}
    \end{tabular}
    \caption{Distribution of labels (CIFAR10)}
    \label{fig:label_dist}
    \centering
\end{figure}

\vspace{2mm}
\noindent\textbf{ML Task and Datasets:} 
We used the benchmark datasets MNIST~\cite{lecun1998gradient} and CIFAR-10~\cite{krizhevsky2009learning}, both 10-class image classification datasets commonly used in FL evaluation. These were partitioned to construct local datasets for each node. We evaluated four scenarios: (i) 3, 5, and 10 nodes, where each node holds samples from 3 or 4 classes; and (ii) a 5-node (uneven) scenario, where the number of classes per node varies. The class distributions are shown in Fig.~\ref{fig:label_dist}. In scenario (i), nodes have roughly equal data amounts, though slight imbalances remain due to label distributions in MNIST. In scenario (ii), nodes with more labels also hold more data.

For MNIST classification, we trained a convolutional neural network (CNN) model. The network consists of a 3×3 valid convolutional layer with 32 channels followed by ReLU, a 3×3 valid convolutional layer with 64 channels and ReLU, a 2×2 max pooling layer, dropout with a rate of 0.25, and a flatten layer. This is followed by a fully connected layer with 128 units, ReLU, dropout with a rate of 0.25, and a final fully connected layer with 10 units and a softmax activation. The model data size \(D_M\) is 38.42\,Mbits, and FLOPs \(M\) is 71.57\,MFLOPs. For CIFAR-10 classification, a pre-trained ResNet18 \cite{he2016deep} was employed, with its output dimension modified to ten. The model data size \(D_M\) is 358.38\,Mbits, and \(M\) is 10.65\,GFLOPs. During FL-based training, all models were optimized using SGD with a learning rate of 0.001 and a momentum of 0.9, and cross-entropy loss was used as the objective function.



\vspace{2mm}
\noindent\textbf{Baselines and Metrics:} As baselines, we compared the following three training scheduling strategies along with the proposed method. Except for the scheduling strategy, all methods follow the Tram-FL training procedure for model training.

\textbf{Random selection} selects one node uniformly at random from $\mathcal{I}$, and the data usage ratio $x_{i,c}^t$ is drawn from a uniform distribution over $[0, 0.1]$.

\textbf{Time-first} aims to minimize the time $\Total{K}$ without considering training data distribution. It selects the node each round yielding the smallest $\Total{K}$, with the data usage fixed at $x_{i,c}^k = 0.1,\ \forall i \in \mathcal I,\ c \in \mathcal C,\ k \in [1,K]$.

\textbf{Variance-first} follows the same strategy as the original Tram-FL, selecting nodes in a fixed cyclic order while ensuring label variance remains below a threshold by controlling the data usage rate. For each selected node, the rate is optimized by solving \eqref{P3}, with the variance constraint set to $V=10000$, which is also used in Load-aware Tram-FL.

As the evaluation metric, we use the time required for the model to achieve a predetermined accuracy. Instead of measuring actual wall-clock time, we employ $\Total{K}$, which accumulates inter-node communication time for model transmission, computation time for training, and waiting time.

\subsection{Result}

Table~\ref{tab:result} presents the total training time $\mathrm{T}_\mathrm{total}^K$ (in seconds) required to reach predefined accuracy thresholds across various node selection strategies, datasets, and federation scenarios. The proposed method, Load-aware Tram-FL, consistently achieves the lowest training time in both MNIST and CIFAR-10 tasks, demonstrating its effectiveness across diverse settings.

In particular, the reduction in training time is especially significant in the 5-node (uneven) scenarios. For example, in the MNIST uneven setting, Load-aware Tram-FL reduces training time by 88\% compared to Random and 85\% compared to Time-first. Similarly, in CIFAR-10, it reduces training time by 72\%, 81\%, and 68\% compared to Random, Time-first, and Variance-first, respectively. This improvement stems from its ability to effectively utilize nodes (e.g., node 0 and node 1) with more IID-like data when resources are available, enabling balanced and efficient training. These results show that Load-aware Tram-FL not only adapts to heterogeneous resource conditions but also implicitly leverages favorable data distributions to enhance training efficiency in non-uniform settings.

\begin{table*}[tbp]
    \centering
    \caption{Comparison of the total training time (in seconds) required to reach predefined accuracy targets under different node selection strategies. Results are shown for MNIST (70\%, 90\%) and CIFAR-10 (60\%, 80\%) across various federation scenarios.}

    \label{tab:result}
    \begin{tabular}{c|c|cccc|cccc}
        \toprule
        \multirow{2}{*}{Method} & \multirow{2}{*}{Scenario} & \multicolumn{4}{c|}{MNIST (Target Acc.: 70\% / 90\%)} & \multicolumn{4}{c}{CIFAR-10 (Target Acc.: 60\% / 80\%)} \\
        \cmidrule(lr){3-6} \cmidrule(lr){7-10}
         & & 3 nodes & 5 nodes & 10 nodes & 5 nodes (uneven) & 3 nodes & 5 nodes & 10 nodes & 5 nodes (uneven) \\
        \midrule
        \multirow{2}{*}{Load-aware Tram-FL}
            & Lower Acc. & 15.69 & \textbf{27.05} & \textbf{16.38} & \textbf{5.09} & \textbf{160.46} & \textbf{150.40} & \textbf{110.44} & \textbf{73.72} \\
            & Higher Acc. & \textbf{31.07} & \textbf{43.13} & \textbf{25.42} & \textbf{11.91} & \textbf{217.44} & \textbf{611.65} & \textbf{273.68} & \textbf{180.00} \\
        \midrule
        \multirow{2}{*}{Random}
            & Lower Acc. & \textbf{12.46} & -- & -- & 42.21 & 226.04 & 483.60 & 523.45 & 259.37 \\
            & Higher Acc. & 46.25 & -- & -- & 85.93 & 330.55 & 1619.09 & 1224.85 & 746.89 \\
        \midrule
        \multirow{2}{*}{Time-first}
            & Lower Acc. & 38.33 & 54.58 & 73.31 & 34.80 & 314.51 & 295.29 & 320.97 & 390.45 \\
            & Higher Acc. & 47.77 & -- & -- & -- & 652.10 & 835.27 & 638.37 & -- \\
        \midrule
        \multirow{2}{*}{Variance-first}
            & Lower Acc. & 24.92 & -- & 106.35 & -- & 306.80 & 842.46 & 550.04 & 232.59 \\
            & Higher Acc. & 64.78 & -- & 171.36 & -- & 608.28 & -- & -- & 961.27 \\
        \bottomrule
    \end{tabular}
\end{table*}

\begin{figure}[tbp]
    \centering
    \subfloat[Test accuracy]{%
        \includegraphics[width=0.47\linewidth]{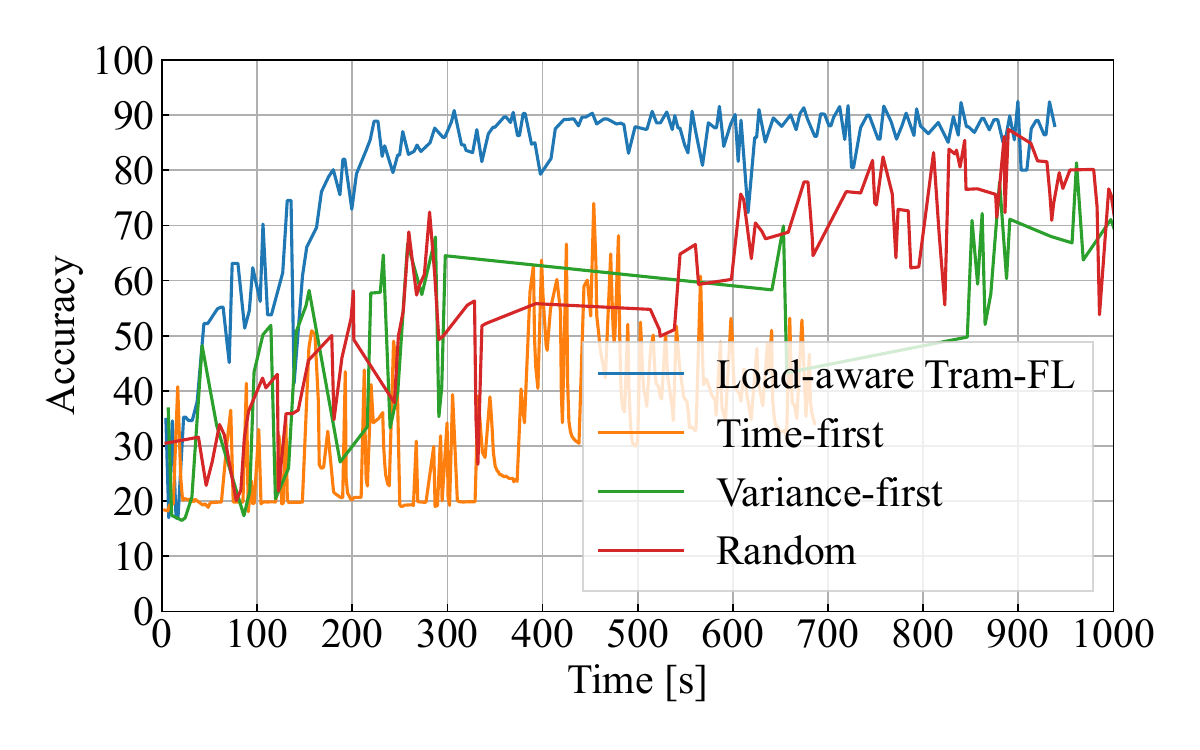}
        \label{fig:accuracy}
    }
    \hfill
    \subfloat[Total training time vs. Round]{%
        \includegraphics[width=0.47\linewidth]{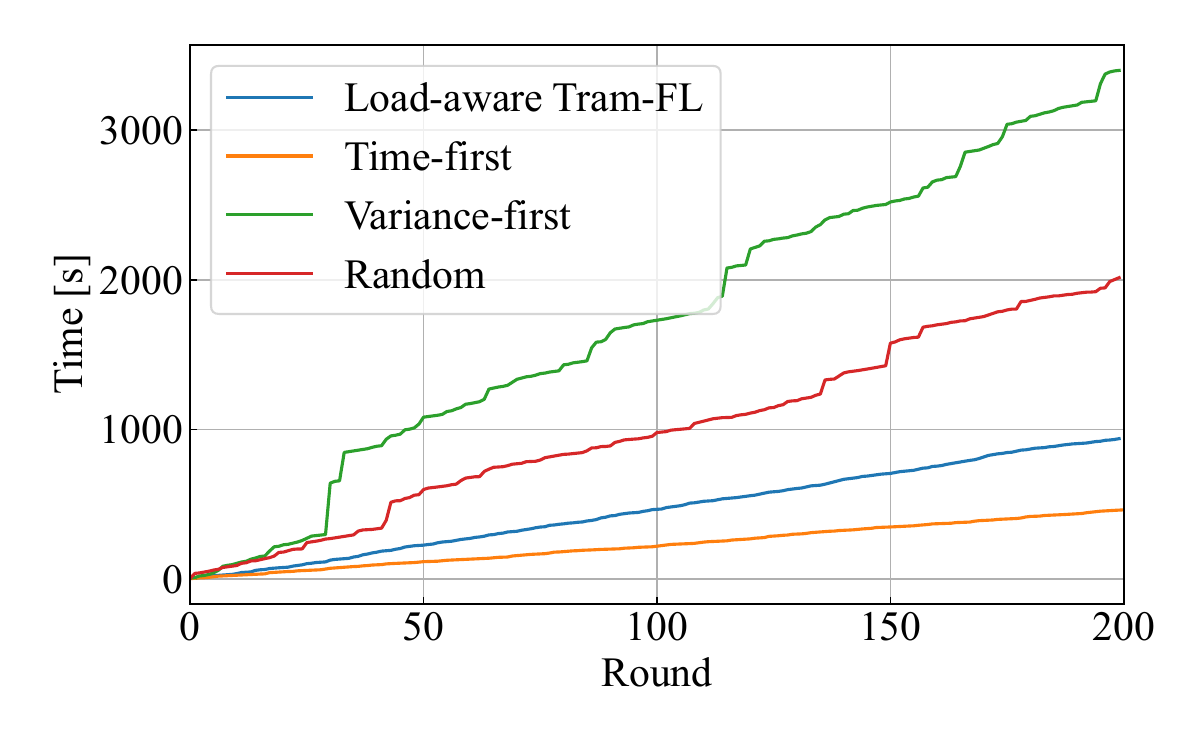}
        \label{fig:time}
    }

    \subfloat[Local validation loss]{%
        \includegraphics[width=0.47\linewidth]{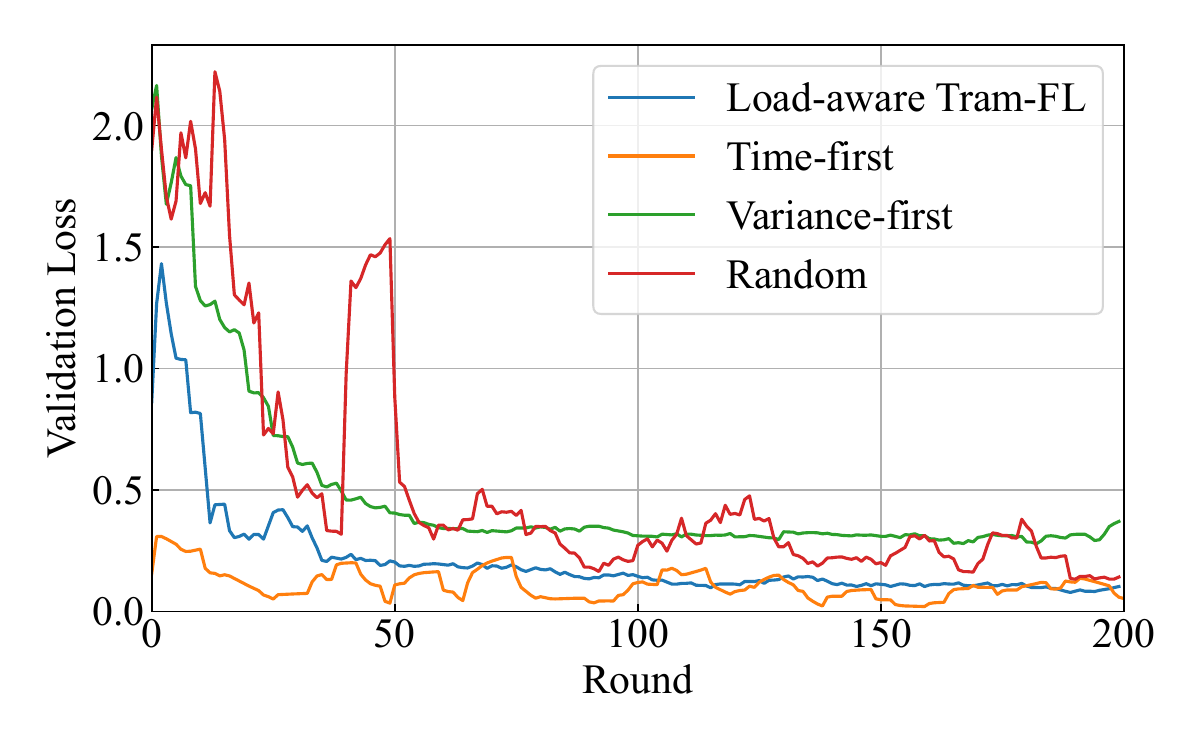}
        \label{fig:local_loss}
    }
    \hfill
    \subfloat[Global test loss]{%
        \includegraphics[width=0.47\linewidth]{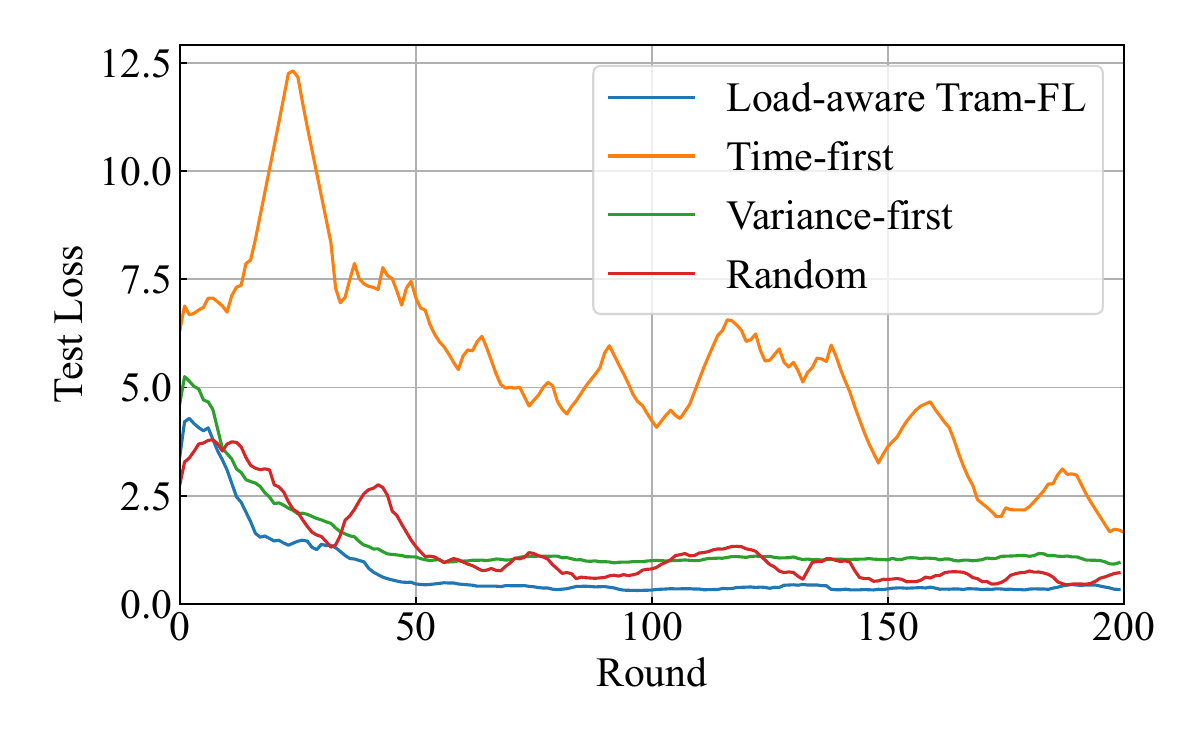}
        \label{fig:global_loss}
    }
    \caption{Training dynamics for CIFAR-10 with 5-node (uneven) scenario.}
    \label{fig:training_dynamics}
\end{figure}

We now focus on the CIFAR-10 5 nodes uneven scenario to closely examine how training progresses under each method. Fig.~\ref{fig:training_dynamics} illustrates the training dynamics of all methods. Note that the x-axis differs across the subplots.

In Fig.~\ref{fig:training_dynamics}(a), which shows how test accuracy evolves over time, the proposed method clearly converges to a higher accuracy faster than the baselines. In contrast, Random and Variance-first exhibit flat regions in their curves, indicating stagnation. This behavior stems from the fact that these methods do not consider the computational or communication load of each node, often resulting in the selection of high-load nodes that significantly delay processing.
Fig.~\ref{fig:training_dynamics}(b) shows the total training time accumulated up to each round. Because the proposed method and Time-first take system load into account when selecting nodes, the training time increases nearly linearly. However, Random and Variance-first exhibit sharp jumps at certain rounds, caused by excessive delays when high-load nodes are selected.

Figs.~\ref{fig:training_dynamics}(c) and (d) show the local validation loss and global test loss per round. Time-first exhibits low local loss but high global loss, indicating overfitting. This is because it minimizes training time without considering data distribution, often selecting the same low-cost node repeatedly. In contrast, the proposed method and Variance-first show stable decreases in both losses, thanks to the variance constraint that promotes balanced training across classes.

\begin{figure}[tbp]
    \centering
    \subfloat[Per-round data usage by node]{%
        \includegraphics[width=0.47\linewidth]{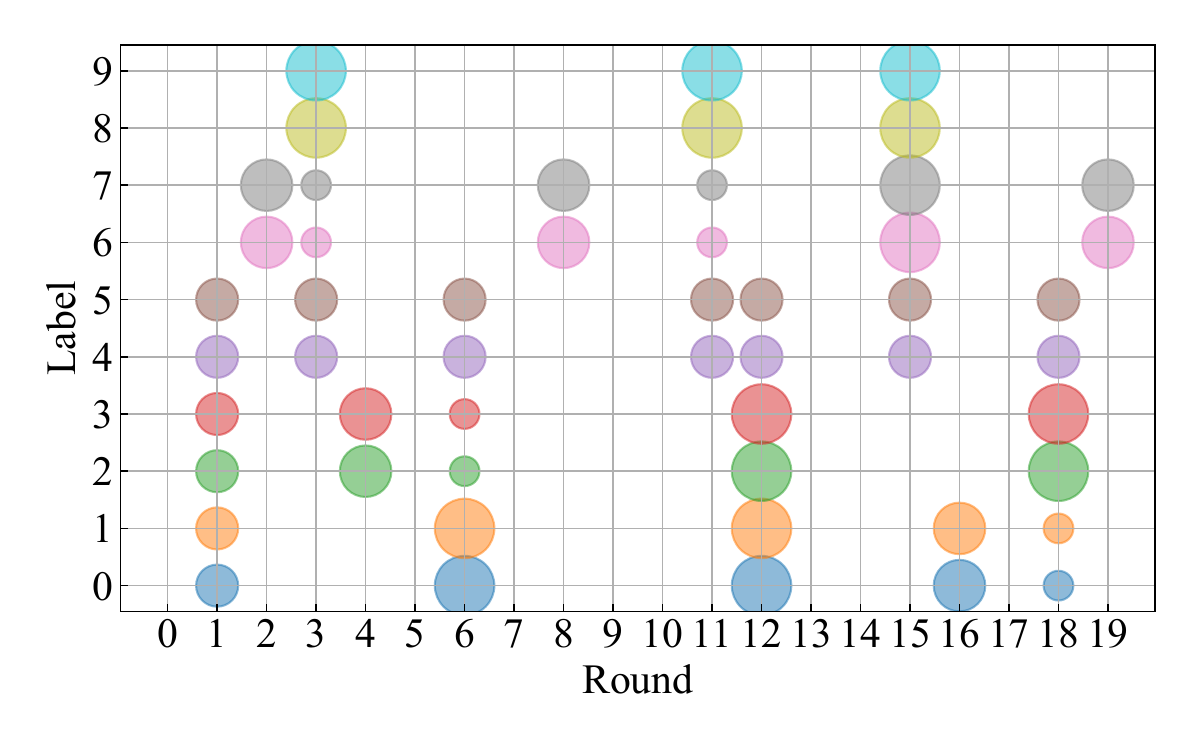}
        \label{fig:data_bubble}
    }
    \hfill
    \subfloat[Cumulative data usage]{%
        \includegraphics[width=0.47\linewidth]{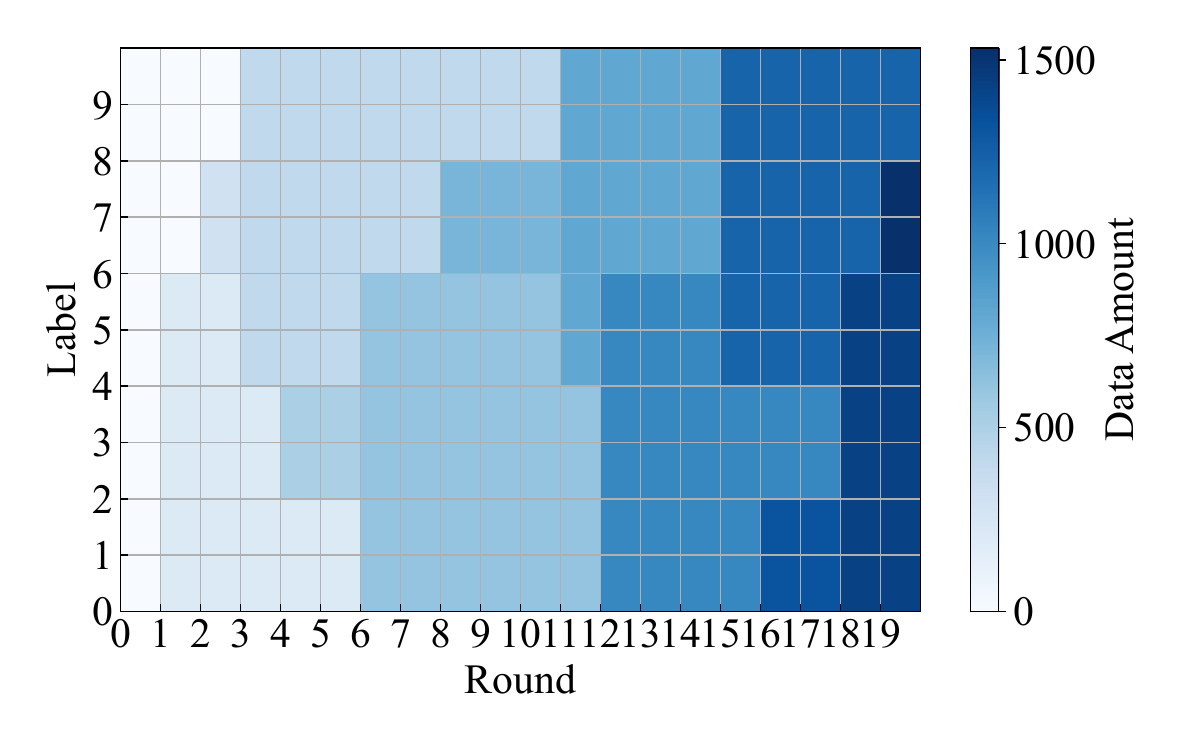}
        \label{fig:data_heatmap}
    }
    \caption{Training data allocation in Load-aware Tram-FL.}
    \label{fig:data}
\end{figure}

Finally, Figs.~\ref{fig:data}(a) and (b) illustrate how the proposed training scheduling promotes balanced use of class data under a non-IID setting. Although each node holds data from only a few classes—causing class imbalance in individual rounds—the cumulative usage becomes increasingly uniform as training progresses. This is due to the variance constraint in the proposed method, which encourages equitable use of all classes and mitigates the negative effects of non-IID data, resulting in faster and more stable model convergence.

%% file: 4_conclusion.tex
\section{Conclusion}

This paper proposes Load-aware Tram-FL, an extension of Tram-FL that introduces a training scheduling method to minimize total training time by accounting for both computational and communication loads. The scheduling problem is formulated as a global optimization task, which—though not directly solvable—is decomposed into node-wise subproblems that are efficiently solvable. Experiments show that the proposed method significantly reduces training time and achieves faster convergence than baselines, even under non-IID data and heterogeneous system conditions. Future work includes handling more dynamic settings where resource availability varies within a round, and designing adaptive algorithms to improve robustness and practicality.
